\documentclass[letterpaper, 10 pt, conference]{ieeeconf}   % paper
\IEEEoverridecommandlockouts     % This command is only
\usepackage{amsmath,amssymb}   % Contains mathematical symbols
\usepackage[T1]{fontenc}
\usepackage{color}
\usepackage{latexsym}
\usepackage[normalem]{ulem}
\usepackage{cite}
\usepackage{graphicx}
\usepackage[english]{babel}    % Language
\usepackage{graphicx}
\usepackage{multirow}
\usepackage{mathtools}
\usepackage{hhline}
\usepackage{siunitx}
\usepackage{dsfont}
\usepackage{pdfpages}
\usepackage{soul}
\usepackage[small]{caption}
\usepackage{url}
\usepackage[labelfont=bf]{caption}
\usepackage[ruled,vlined]{algorithm2e}
\newtheorem{theorem}{Theorem}

\newtheorem{lemma}[theorem]{Lemma}
\newtheorem{definition}{Definition}

\newtheorem{problem}{Problem}

\title{Online Planning of Uncertain MDPs under Temporal Tasks\\
 and Safe-Return Constraints}
\author{Yuyang Zhang and Meng Guo
  \thanks{The authors are with the College of Engineering, Peking University,
Beijing 100871, China.
    E-mail: \texttt{yuyangzhang@stu.pku.edu.cn, meng.guo@pku.edu.cn}.
    This work is supported by the Natural Science Foundation of China (NSFC)
    under Grant-62203017.}
}

\begin{document}
\maketitle \thispagestyle{empty} \pagestyle{empty}
%=============================
%=============================
\begin{abstract}
  This paper addresses the online motion planning problem of
  mobile robots under complex high-level tasks.
  The robot motion is modeled as an uncertain Markov
  Decision Process (MDP) due to limited initial knowledge,
  while the task is specified as Linear Temporal Logic (LTL) formulas.
  The proposed framework enables the robot to explore and update the
  system model in a Bayesian way,
  while simultaneously optimizing the asymptotic costs of satisfying the
  complex temporal task.
  Theoretical guarantees are provided for the synthesized outgoing policy
  and safety policy.
  More importantly, instead of greedy exploration
  under the classic ergodicity assumption,
  a safe-return requirement is enforced such that the robot can always
  return to home states with a high probability.
  The overall methods are validated by numerical simulations.
\end{abstract}
%==============
%=============================
\section{Introduction}\label{sec:introduction}
Uncertainty arises in various aspects of robot motion planning such as
the model of the workspace and the outcome of motion execution.
Markov Decision Process (MDP) is a convenient way to model such
uncertain systems~\cite{davis2018markov}
based on which decision making problems are solved to optimize
a given control objective.
The most common objective is to reach a set of goal states
while minimizing the expected total cost.
The resulting solution is a policy that maps states to
probability distributions over the set of
allowed actions~\cite{davis2018markov}.
%LTL with MDP
Furthermore, there have been many efforts to address the problem
of synthesizing a control policy for a MDP that satisfies
high-level temporal tasks.
Most common control objectives such as reachability, surveillance,
liveness and emergency response, can be specified via
temporal logic formulas.
There has been numerous work considering different formal languages,
such as Probabilistic Computation Tree Logic (PCTL)
and Linear Temporal Logics (LTL), see\cite{ding2014optimal}.
Such tasks are normally specified over regions of interest
in the state space.
A verification toolbox is provided in~\cite{etessami2007multi} for MDPs
under certain LTL tasks.
Different cost optimizations are also considered such as maximum
reachability in~\cite{baier2008principles},
the minimal bottleneck cost in~\cite{ding2014optimal},
the pareto resource constraints in~\cite{bruyere2016meet},
the balanced satisfiability and cost in \cite{guo2018probabilistic},
and the uncertainty over semantic maps
in~\cite{kantaros2022perception}.

%learning + safety
However, under limited information,
even the underlying MDP could be uncertain,
e.g., the transition measure or the state features is only partially-known,
the above techniques can not be applied directly.
Thus,  robust control policies are synthesized offline
in~\cite{ding2014ltl} to maximize the accumulated time-varying rewards,
in~\cite{ahmed2017sampling} to maximize the satisfiability under
uncertain transition measures, and in~\cite{guo2015multi} to improve
multi-robot team performance for dynamic workspaces.
These policies are mostly constructed offline.
In contrast, online approaches require the robot to actively explore
and learn the system model and the optimal policy simultaneously
during run time.
The work in~\cite{qian2019exploration} introduces
{exploration bonus} to balance exploration and exploitation
during learning. To guarantee convergence, most existing
exploration algorithms rely on the assumption of \emph{ergodicity} that any
state in the MDP is reachable from any other state under a suitable
policy~\cite{talebi2018variance}.
Thus, any state can be safely explored and consequently
the system model around that state.
Nonetheless, this assumption does not hold in many practical examples
where the system would \emph{break} once entering an unsafe state,
e.g., a ground vehicle falls off stairs, or enters a room via a one-way door.
Thus, safe exploration during learning has been an active
research topic, see~\cite{garcia2015comprehensive}.
Nonetheless, complex temporal tasks have not been well studied within
non-ergodic systems that are partially-unknown.

To overcome these issues, this work proposes an online planning and exploration
method for robotic systems modeled as uncertain MDPs. It allows the robot to
gradually improve the model and thus the asymptotic cost of the complex task,
while ensuring that it can always safely return to a set of home states.
The main contribution lies in the novel framework
for general uncertain MDPs, which can handle complex temporal tasks
and ensure real-time safety during the learning processes.

%======================
%======================
\section{Preliminaries}\label{sec:prelims}

\subsection{Linear Temporal Logic (LTL)}\label{subsec:LTL}
The ingredients of a LTL formula are
a set of atomic propositions $AP$ and several Boolean and
temporal operators. Atomic propositions are Boolean variables
that can be either true or false.  A LTL formula is specified
according to the syntax~\cite{baier2008principles}:
$\varphi \triangleq \top \;|\; p  \;|\;
\varphi_1 \wedge \varphi_2  \;|\;
\neg \varphi  \;|\;\bigcirc \varphi  \;|\;
\varphi_1 \,\textsf{U}\, \varphi_2,$
where $\top\triangleq \texttt{True}$, $p \in AP$,
$\bigcirc$ (\emph{next}), $\textsf{U}$ (\emph{until})
and $\bot\triangleq \neg \top$.
We omit the derivations of other operators like $\square$
(\emph{always}), $\Diamond$ (\emph{eventually}),
$\Rightarrow$ (\emph{implication}).
Given any word $w$ over~$AP$, it can be verified whether $w$ satisfies the formula,
denoted by~$w\models \varphi$.
The full semantics and syntax of LTL are omitted here,
see e.g.,~\cite{baier2008principles}.

\subsection{Deterministic Rabin Automaton~(DRA)}\label{subsec:dra}
The set of words that satisfy a LTL formula~$\varphi$ over $AP$ can
be captured through a Deterministic Rabin
Automaton~(DRA)~$\mathcal{A}_{\varphi}$~\cite{baier2008principles},
defined as~$\mathcal{A}_{\varphi}\triangleq (Q, \,2^{AP},\, \delta,
\, q_0,\,\text{Acc}_{\mathcal{A}})$,
where $Q$ is a  set of states; {$2^{AP}$ is the alphabet};
$\delta\subseteq Q\times 2^{AP} \times {Q}$ is a transition relation;
$q_0 \in Q$ is the initial state;
and~$\text{Acc}_{\mathcal{A}} \subseteq 2^Q \times 2^Q$ is
a set of accepting pairs, i.e., $\text{Acc}_{\mathcal{A}} =
\{(H^1_{\mathcal{A}}, I^1_{\mathcal{A}}),
(H^2_{\mathcal{A}}, I^2_{\mathcal{A}}), \cdots,
(H^N_{\mathcal{A}}, I^N_{\mathcal{A}})\}$
where~$H^i_{\mathcal{A}},\, I^i_{\mathcal{A}}\subseteq Q$,
$\forall i =1,2,\cdots, N$.
An infinite run~$q_0q_1q_2\cdots$ of~$\mathcal{A}$ is \emph{accepting}
if there exists \emph{at least one} pair~$(H^i_{\mathcal{A}},
\, I^i_{\mathcal{A}})\in \text{Acc}_{\mathcal{A}}$ such
that~$\exists n\geq 0$, it holds~$\forall m\geq n,\,
q_m\notin H^i_{\mathcal{A}}$ and~$\overset{\infty}{\exists}
k\geq0$,~$q_k\in I^i_{\mathcal{A}}$, where~$\overset{\infty}{\exists}$
stands for~``existing infinitely many''.
Namely, this run should intersect with~$H^i_{\mathcal{A}}$
\emph{finitely} many times while
with~$I^i_{\mathcal{A}}$ \emph{infinitely} many times.
There are translation tools~\cite{klein2007ltl2dstar}
to obtain~$\mathcal{A}_{\varphi}$ given~$\varphi$ with
complexity~$2^{2^{\mathcal{O}(|\varphi|\log |\varphi|)}}$.

%==============================================================
\section{Problem Formulation}\label{sec:problem-formulate}
\subsection{Probabilistically-labeled MDP}
We extended the probabilistically-labeled MDP proposed in our
earlier work~\cite{guo2018probabilistic} to include uncertainty
in robot motion and workspace properties:
\begin{equation}\label{eq:mdp}
  \mathcal{M} \triangleq (X, \, U,\, D,\, p_{D}, \, (x_0,\,l_0),
  \, AP, \,L,\, p_{L},\,c_D),
\end{equation}
where~$X$ is the finite state space;
$U$ is the finite control action space
and $U(x)$ denotes the set of actions \emph{allowed}
at state~$x\in X$;
$D \triangleq \{(x,u)\,|\, x\in X,\, u\in U(x)\}$ is the set
of possible state-action pairs;
$p_D\colon X\times U \times X \rightarrow {[0, 1]}$ is the
transition probability
and~$\sum_{\check{x}\in X}p_D(x,u,\check{x}) = 1$,
$\forall (x,\,u) \in D$;
$c_D \colon D \rightarrow \mathbb{R}^{>0}$ is the cost function;
$AP$ is a set of atomic propositions as the properties of interest;
~$L \colon X \rightarrow 2^{2^{AP}}$ returns the properties held
at each state; and~$p_{L} \colon X\times 2^{AP}\rightarrow {[0,\,1]}$ is
the associated probability. Note that~$\sum_{l \in L(x)} p_L(x,\,l)=1$,
$\forall x \in X$;and~$x_0\in X$,~$l_0\in L(x_0)$ are the initial states and labels.

\subsection{Uncertainty and Bayesian Learning}\label{subsec:uncertain}

However, due to limited initial knowledge,
the above MDP is \emph{uncertain}. Particularly,
for each pair $(x,u)\in D$, the distribution over its post states
follows the Dirichlet distribution~\cite{mackay2003information}
with parameter $\boldsymbol{\alpha}_x^u$:
\begin{equation}\label{eq:dirichlet-D}
p_D \sim \text{Dirichlet}(\boldsymbol{b}_x^u,\, \boldsymbol{\alpha}_x^u),
\end{equation}
where $\boldsymbol{b}_x^u\triangleq\{b_0,b_1,\cdots,b_{K_x^u}\}$,
where $b_k$ is a vector of length $K_x^u$ with one
at index $k$ and the remaining elements are zero,
$\forall k=0,\cdots,K_x^u$;
$\boldsymbol{\alpha}_x^u\triangleq\{\alpha_{x}^u(\check{x}),
\forall \check{x}\in \mathcal{K}_x^u\}$ is a set of
non-negative scaling coefficients with
$\alpha_{x}^u(\check{x})\geq 0$;
also~$\mathcal{K}_x^u\triangleq|\check{x}\in X\, |\, p_D(x,u,\check{x})>0|$,
and~$K_x^u\triangleq|\mathcal{K}_x^u|$.

Similarly, for each $x \in X$, the distribution over its labels
also follows the Dirichlet distribution with parameter $\boldsymbol{\alpha}_x^L$:
\begin{equation}\label{eq:dirichlet-L}
p_L \sim \text{Dirichlet}(\boldsymbol{b}_x^L,\, \boldsymbol{\alpha}_x^L),
\end{equation}
where $\boldsymbol{b}_x^L\triangleq\{b_0,b_1,\cdots,b_{K_x^L}\}$ is defined
similarly to $\boldsymbol{b}_x^u$;
$\boldsymbol{\alpha}_x^L\triangleq\{\alpha_{x}^L(l),\forall l\in L(x)\}$
is a set of non-negative scaling coefficients $\alpha_{x}^L(l)\geq 0$;
and $K_x^L \triangleq |l\in L(x)\, |\, p_L(x,l)>0|$.
Thus, we denote the complete set of parameters that govern the transition
and labeling probability of~$\mathcal{M}$ by:
\begin{equation}\label{eq:alpha}
  \boldsymbol{\alpha}\triangleq\{\boldsymbol{\alpha}_{x}^u,\,
  \boldsymbol{\alpha}_x^L, \,\forall (x,u)\in D\},
\end{equation}
which is called the \emph{belief} over $\mathcal{M}$.
In the sequel, we use~$\mathcal{M}^{\boldsymbol{\alpha}}$ to denote the
general class of MDP $\mathcal{M}$ under belief $\boldsymbol{\alpha}$,
while $\mathcal{M}$ alone stands for one sample
from $\mathcal{M}^{\boldsymbol{\alpha}}$.

Furthermore, the robot is equipped with sensors and thus can observe
the actual transitions and labels during motion.
Then, the distributions~$p_D$ and~$p_L$ can be updated in a Bayesian way
by following~\cite{neal2000markov}.

\subsection{Task Specification}
Moreover, there is a LTL task formula~$\varphi$ specified over
the same set of atomic propositions~$AP$ as the desired
behavior of~$\mathcal{M}$, following the syntax in Sec.~\ref{subsec:LTL}.

At stage~$T\geq 0$, the robot's past path is given
by~$X_T= x_0 x_1 \cdots x_T \in X^{(T+1)}$,
the past sequence of observed labels is given by~$L_T=l_0
l_{1}\cdots l_T\in (2^{AP})^{(T+1)}$ and the past sequence of
control actions is~$U_T=u_0 u_1 \cdots u_T \in U^{(T+1)}$.
It should hold that~$p_D(x_t, u_t, x_{t+1})>0$ and~$p_L(x_t, l_t)>0$,
$\forall t\geq 0$. The complete past is then given by~$R_T=x_0l_0u_0
\cdots x_Tl_Tu_T$. Denote by~$\boldsymbol{X}_T$, $\boldsymbol{L}_T$
and~$\mathbf{R}_T$ the set of all possible past sequences of states,
labels, and runs up to stage~$T$. We set $T=\infty$ for infinite sequences.
Then, the \emph{mean} total cost~\cite{davis2018markov} of an infinite
robot run~$R_{\infty}$ of~$\mathcal{M}$ is defined
as~$\textbf{Cost}(R_{\infty}) \triangleq \lim_{t\rightarrow \infty}
\frac{1}{t}\sum_{t=0}^{\infty} \,c_D(x_t,u_t)$,
where $c_D(\cdot)$ is the cost of applying $u_t$ and $x_t$
from~\eqref{eq:mdp}. A \emph{finite-memory} policy is defined
as~$\boldsymbol{\mu}=\mu_0\mu_1\cdots$. The control policy at
stage~$t\geq 0$ is given by~$\mu_t:\mathbf{R}_t \times U
\rightarrow [0, 1]$, $\forall t\geq 0$.
Denote by~$\overline{\boldsymbol{\mu}}$ the set of all such
finite-memory policies.

Given one sample MDP $\mathcal{M}$ and a policy $\boldsymbol{\mu}$,
the set of all infinite runs is denoted by
$\mathbf{R}_{\mathcal{M}}^{\boldsymbol{\mu}}\subset \mathbf{R}_{\infty}$.
Then the probability of~$\mathcal{M}$ satisfying~$\varphi$
under ~$\boldsymbol{\mu}$  is defined by:
\begin{equation}\label{eq:satisfy}
  \textbf{Sat}_{\mathcal{M}}^{\boldsymbol{\mu}}\triangleq
  Pr_{{\mathcal{M}}}^{\boldsymbol{\mu}}(\varphi)={Pr}_{\mathcal{M}}^{\boldsymbol{\mu}}
  \big{(} \mathbf{R}_{\mathcal{M}}^{\boldsymbol{\mu}} \,|\,
  \boldsymbol{L}_{\infty}\models \varphi \big{)},
\end{equation}
where the satisfaction relation~``$\models$'' is introduced in
Sec.~\ref{subsec:LTL}. Namely, the~\emph{satisfiability} equals
to the probability of all infinite runs whose associated labels
satisfy the task. More details on the probability measure can be
found in~\cite{baier2008principles}. Moreover, the \emph{cost} of
policy $\boldsymbol{\mu}$ over~$\mathcal{M}$ is denoted by
\begin{equation}\label{eq:total-cost}
\textbf{Cost}_{\mathcal{M}}^{\boldsymbol{\mu}}\triangleq \mathbb{E}_{R_\infty\in \mathbf{R}_{\mathcal{M}}^{\boldsymbol{\mu}}}\{\textbf{Cost}(R_\infty)\},
\end{equation}
 as the expected mean cost of all possible infinite runs.

%%%%%%%%%%
\subsection{Safe-return Constraints}\label{subsec:safety}
Furthermore, to ensure safety while the robot explores the workspace,
we introduce the following definition of safety based on~\cite{moldovan2012safe}.
Particularly, consider two finite-memory policies $\boldsymbol{\mu}_{\texttt{o}},
\boldsymbol{\mu}_{\texttt{r}}\in \overline{\boldsymbol{\mu}}$,
where $\boldsymbol{\mu}_{\texttt{o}}$ is called the \emph{outbound}
policy that drives the robot to satisfy task~$\varphi$
and~$\boldsymbol{\mu}_{\texttt{r}}$ is the \emph{return} policy that ensures
the safety constraint below.

\begin{definition}\label{def:safety}
  Given system $\mathcal{M}$, an outbound policy~$\boldsymbol{\mu}_{\texttt{o}}$
  is called $\chi_{\texttt{r}}$-\emph{safe} at stage $t\geq 0$
  if there exists a \emph{return} policy~$\boldsymbol{\mu}_{\texttt{r}}$
  such that the probability of system $\mathcal{M}$ returning
  to a set of home states~$X_{\texttt{r}}\in X$ is lower-bounded, namely,
\begin{equation}\label{eq:return}
  \textbf{Safe}_{\mathcal{M}}^{\boldsymbol{\mu}_{\texttt{o}},\boldsymbol{\mu}_{\texttt{r}}}
  \triangleq Pr_{{\mathcal{M}},(x_t,l_t)}^{\boldsymbol{\mu}_{\texttt{o}},
    \boldsymbol{\mu}_{\texttt{r}}}(\Diamond X_{\texttt{r}}) \geq \chi_{\texttt{r}},
\end{equation}
where $\chi_{\texttt{r}}>0$ is the assigned safety bound,
$(x_t,l_t)$ are the robot state and label at stage $t$. \hfill $\blacksquare$
\end{definition}

Note that traditionally safety is defined as the avoidance of a set
of \emph{unsafe} states,
see~\cite{garcia2015comprehensive},
which mostly are policy-independent and given before-hand.
Despite its intuitiveness,  it has serious drawbacks in scenarios
where unsafe states can only be determined during run time, thus
unknown beforehand. In contrast, the safety measure in~\eqref{eq:return}
is policy-dependent and can cover the traditional notion.

\subsection{Problem Statement}\label{subsec:problem}

\begin{problem}\label{prob:main}
  Given the class of uncertain MDPs~$\mathcal{M}^{\boldsymbol{\alpha}}$
  from~\eqref{eq:mdp}-\eqref{eq:dirichlet-L}, and the task
  specification~$\varphi$, our goal is to synthesize the outbound and
  return polices $\boldsymbol{\mu}_{\texttt{o}},\boldsymbol{\mu}_{\texttt{r}}$
  at \emph{each} stage $t\geq 0$ that solve the constrained optimization below:
\begin{equation}\label{eq:objective}
\begin{split}
  &\min_{\boldsymbol{\mu}_{\texttt{o}},\boldsymbol{\mu}_{\texttt{r}}
    \in \overline{\boldsymbol{\mu}}}
  \;\; \mathbb{E}_{\boldsymbol{\alpha}}
  \{\textbf{Cost}_{\mathcal{M}}^{\boldsymbol{\mu}_{\texttt{o}}}\}\\
  &\; \text{s.t.} \quad \mathbb{E}_{\boldsymbol{\alpha}}
  \{\textbf{Sat}_{\mathcal{M}}^{\boldsymbol{\mu}_{\texttt{o}}}\}
  \geq \chi_{\texttt{o}}\;\; \text{and} \;\;
  \mathbb{E}_{\boldsymbol{\alpha}}\{
  \textbf{Safe}_{\mathcal{M}}^{\boldsymbol{\mu}_{\texttt{o}},
    \boldsymbol{\mu}_{\texttt{r}}}\}\geq \chi_{\texttt{r}},
\end{split}
\end{equation}
where~$\boldsymbol{\alpha}$ is the belief from~\eqref{eq:alpha},
and $\chi_{\texttt{o}},\chi_{\texttt{r}}>0$ are given
lower bounds for satisfiability and safety in~\eqref{eq:satisfy}
and~\eqref{eq:return}.
The expectation over alpha before the
$\textbf{Sat}_{\mathcal{M}}^{\boldsymbol{\mu}_{\texttt{o}}}$
and
$\textbf{Safe}_{\mathcal{M}}^{\boldsymbol{\mu}_{\texttt{o}},
    \boldsymbol{\mu}_{\texttt{r}}}$
is due to the uncertainty in the system model
$\mathcal{M}^{\boldsymbol{\alpha}}$.
\hfill $\blacksquare$
\end{problem}

Main difficulty of the above problem comes from the uncertainties
in $\mathcal{M}$ and the consideration of complex temporal tasks
along with policy-dependent safety constraints.

It is worth noting that
the safe-return constraint in~\eqref{eq:return} can \emph{not} be
  treated as an additional task of the original task~$\varphi$,
  as they are surely conflicting objectives.
  Thus, the methods proposed in~\cite{bruyere2016meet,
guo2018probabilistic} that synthesize only
  one policy~$\boldsymbol{\mu}$ to satisfy simultaneously both
  tasks, can not be applied here.
  In other words, it is essential to synthesize \textbf{two}
  polices: $\boldsymbol{\mu}_{\texttt{o}}$ for the actual task and
  $\boldsymbol{\mu}_{\texttt{r}}$ for the safe-return requirement.

%------------------------------
%========================================
\begin{figure}[t]
    \centering
    \includegraphics[width=0.99\linewidth]{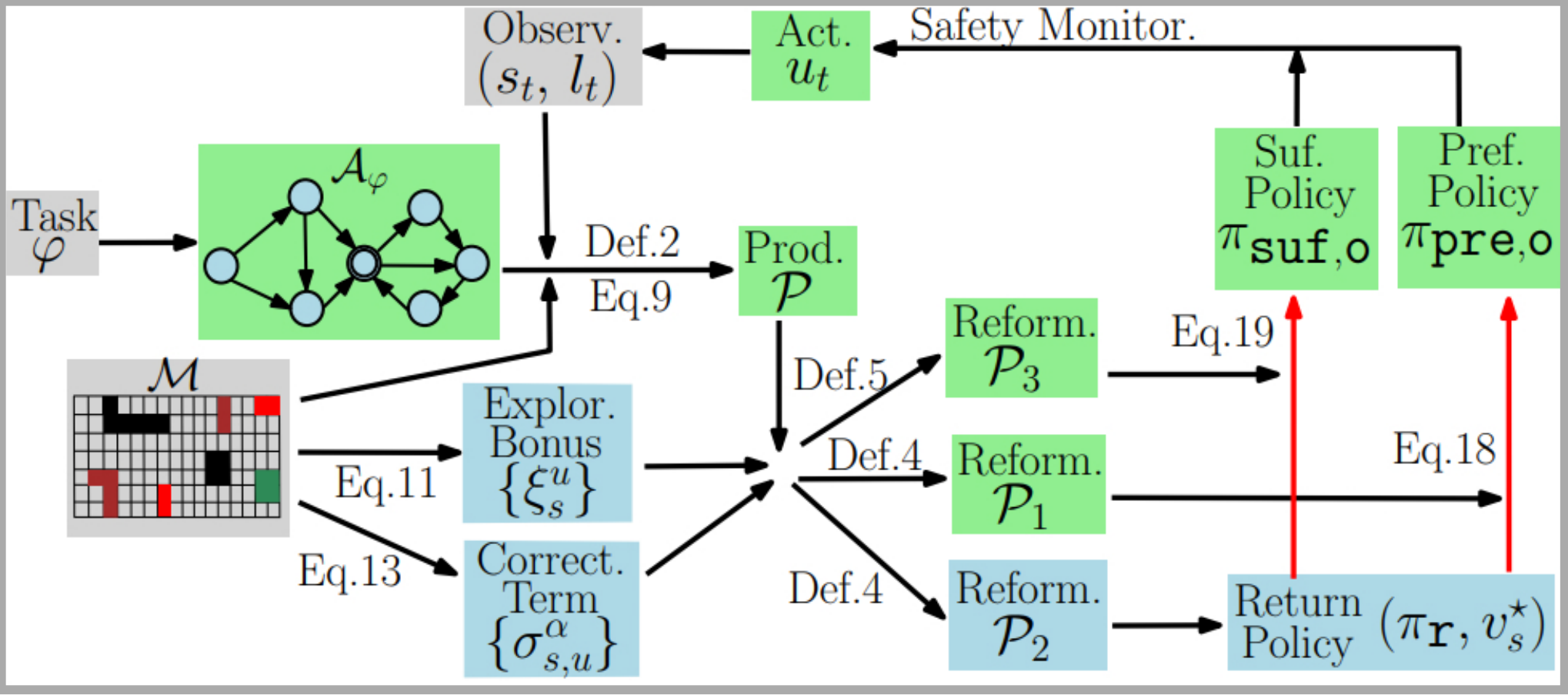}
    %--------------------
    \caption{Illustration of the proposed framework.}
    \label{fig:framework}
    %--------------------
    \vspace{-0.15cm}
\end{figure}
%========================================

%=========================================================
%=========================================================
\section{Safety and Task Policy Synthesis}\label{sec:synthesis}
In this section, we describe the key steps to synthesize
the safety and task policies.
As shown in Fig.~\ref{fig:framework}, both policies
are used in the online execution described in the sequel.

%=========================================================
\subsection{Product Automaton and AMECs}\label{subsec:product}
To begin with, we construct the DRA~$\mathcal{A}_{\varphi}$ associated
with the~LTL task formula~$\varphi$ via the translation
tools~\cite{klein2007ltl2dstar}.
Let it be~$\mathcal{A}_{\varphi}=(Q, \,2^{AP},\, \delta,\,
q_{0},\,\text{Acc}_{\mathcal{A}})$, where the notations are
defined in Sec.~\ref{subsec:dra}.
Then we construct a product automaton between the
model~$\mathcal{M}$ and the DRA~$\mathcal{A}_{\varphi}$.

\begin{definition}\label{def:product}
  The product $\mathcal{P}\triangleq \mathcal{M}\times \mathcal{A}_{\varphi}$
  is a 7-tuple:
\begin{equation}\label{eq:prod}
\mathcal{P}=(S,\,U,\,E,\,p_E,\,c_E,\, s_0,\,\text{Acc}_{\mathcal{P}}),
\end{equation}
where: the state~$S\subseteq X\times 2^{AP} \times Q$
satisfies~$\langle x,\,l,\,q \rangle \in S$,
$\forall x \in X$,~$\forall l\in L(x)$ and~$\forall q\in Q$;
the action set~$U$ is the same as in~\eqref{eq:mdp}
and~$U(s)=U(x)$,~$\forall s=\langle x,l,q\rangle \in S$;
$E=\{(s,u)\,|\, s\in S,\, u\in U(s)\}$;
the transition probability~$p_E \colon S\times U \times
S\rightarrow {[0,\,1]}$ is defined by
\begin{equation}\label{eq:new_pro}
  p_E\big{(}\langle x,l,q\rangle,\,u,\, \langle \check{x},
  \check{l},\check{q}\rangle \big{)} =  p_D(x,\,u,\,\check{x})
  \cdot p_L(\check{x},\,\check{l})
\end{equation}
where (i) $\langle x,l,q\rangle,\, \langle \check{x},\check{l},
\check{q}\rangle \in S$;
(ii)~$(x,u)\in D$; and
(iii)~$\check{q}= \delta(q,\,l)$;  the cost function~$c_E \colon
E\rightarrow \mathbb{R}^{>0}$ is given by~$c_E\big{(}
\langle x,l,q\rangle,\,u \big{)}=c_D(x,u)$, $\forall
\big{(}\langle x,l,q\rangle,\,u \big{)}\in E$. Namely,
the label~$l$ should fulfill the transition condition
from~$q$ to~$\check{q}$ in~$\mathcal{A}_{\varphi}$; the
single initial state is~$s_0=\langle x_0,l_0,q_0 \rangle
\in S$; the accepting pairs are defined
as~$\text{Acc}_{\mathcal{P}}=\{(H^i_{\mathcal{P}},\, I^i_{\mathcal{P}}),
i=1,\cdots,N\}$, where~$H^i_{\mathcal{P}}=\{\langle x,l,q\rangle
\in S\,|\, q\in H^i_{\mathcal{A}}\}$ and~$I^i_{\mathcal{P}}
=\{\langle x,l,q\rangle \in S\,|\, q\in I^i_{\mathcal{A}}\}$,
~$\forall i=1,\cdots,N$. \hfill $\blacksquare$
\end{definition}

The product~$\mathcal{P}$ computes the
intersection between the traces of~$\mathcal{M}$ and the words
of~$\mathcal{A}_{\varphi}$, to find the admissible robot behaviors
that satisfy the task~$\varphi$. It combines the uncertainty in
robot motion and the workspace model by including both~$x$ and~$l$
in the states. For simpler notation,
let $\mathcal{K}_s^u=\{\check{s}\in S\, |\, p_E(s,u,\check{s})>0\}$
and $K_s^u=|\mathcal{K}_s^u|$. Note that since $\mathcal{M}$ is
uncertain under belief $\boldsymbol{\alpha}$, we denote by
$\mathcal{P}^{\boldsymbol{\alpha}}$ the general class of product
automata associated with each sample MDP
within~$\mathcal{M}^{\boldsymbol{\alpha}}$.
Lastly, the set of home states in~$\mathcal{P}$ is
denoted by $S_{\texttt{r}}\triangleq~\{s\in S\,|\,s=
\langle x,l,q\rangle, x\in X_{\texttt{r}}\}$.

The accepting condition of~$\mathcal{P}$ is the same as
in Sec.~\ref{subsec:dra}. To transform this condition into
equivalent graph properties,  we first compute the
accepting maximum end components (AMECs) of~$\mathcal{P}$
associated with its accepting pairs~$\text{Acc}_{\mathcal{P}}$.
Denote by~$\Xi_{acc}=\{(S'_1,\, U'_1), (S'_2,\, U'_2),\cdots (S'_{C},\, U'_{C})\}$
the set of AMECs associated with~$\text{Acc}_{\mathcal{P}}$,
where~$S'_{c}\subset S$ and~$U'_c:S'_c \rightarrow 2^{U}$,~$\forall c=1,2,
\cdots,C$.
Note that $S'_{c_1}\cap S'_{c_2}=\emptyset$, $\forall c_1,c_2=1,\cdots,C$.
We omit the definition and derivation of~$\Xi_{acc}$ here,
and refer the readers to
Definition~10.124 of~\cite{baier2008principles}.

%=========================================================
\subsection{Safe Exploration and Policy Synthesis}\label{subsec:explore}
In this part, we explain how to introduce exploration bonus to
encourage exploration in addition to the tasks.
More importantly, we formally prove how the
safety and satisfiability constraints under uncertain MDPs
can be re-formulated as the policy synthesis
under standard MDPs.

\subsubsection{Exploration Bonus}\label{subsubsec:bonus}
The notion of exploration bonus has been proposed to encourage
exploration during the policy learning. Intuitively, this
approach would drive the system to try state-action pairs
that have \emph{not} been observed enough times by
assuming a high bonus there.

\begin{definition}\label{def:bonus}
  Given the pair $(s,\, u)\in E$ in $\mathcal{P}$,
  where $s=\langle x,l,q \rangle$ and the associated
  Dirichlet parameters $\boldsymbol{\alpha}_x^u$,
  $\boldsymbol{\alpha}_x^L$, the exploration bonus
  of choosing action $u$ at state $s$, denoted
  by~$\xi_s^u\in \mathbb{R}^+$, is defined by:
\begin{equation}\label{eq:bonus}
\xi_s^u = \begin{dcases}
  0, \qquad \text{if } \overline{{\alpha}}_x^u>
  \alpha_U \text{ and } \hspace{-0.1in} &
  \overline{{\alpha}}_x^L>\alpha_L;  \\
  \frac{g_U}{1+\overline{{\alpha}}_x^u} +
  \frac{g_L}{1+\overline{{\alpha}}_x^L}, & \text{otherwise};
\end{dcases}
\end{equation}
where  $\overline{{\alpha}}_x^u\triangleq \sum_{\check{x}\in \mathcal{K}_x^u}
\alpha_x^u(\check{x})$, $\overline{{\alpha}}_x^L\triangleq
\sum_{\check{x}\in \mathcal{K}_x^u} \sum_{l\in L(\check{x})}\alpha_{\check{x}}^L(l)$;
and $g_U, g_L, \alpha_U, \alpha_L>0$ are pre-defined constants.
\hfill $\blacksquare$
\end{definition}

In other words, the more a state $x$ has been visited and
an action $u$ is chosen at state $x$, the less the exploration
bonus $\xi_s^u$ is.
The MDP $\mathcal{M}$ is called \emph{fully explored}
if the first case of~\eqref{eq:bonus} holds for all $(s,\, u)\in E$.

\subsubsection{Constraints Reformulation}\label{subsubsec:reformulation}

As proven in Theorem~1 of~\cite{moldovan2012safe},
it is in general \emph{NP-hard} to decide whether
there exists a $\chi_{\texttt{r}}$-safe policies for a
given MDP~$\mathcal{P}$ under belief~$\boldsymbol{\alpha}$,
except only very limited cases. Thus, we rely on the following
two theorems to reformulate the satisfiability and
safety constraints in Problem~\ref{prob:main}.

\begin{definition}\label{def:MDPs}
  Consider two variants of MDP~$\mathcal{P}$: the first
  MDP~$\mathcal{P}_1\triangleq (1-b_{s,S_{\Xi}})\cdot{\mathcal{P}}$
  where $b_{s,S_{\Xi}}\triangleq \mathds{1}_{\{s \in S_{\Xi}\}}$ is an
  indicator function.
  The second MDP~$\mathcal{P}_2\triangleq (1-b_{s,S_{\texttt{r}}})
  \cdot{\mathcal{P}}$, where $b_{s,s_{\texttt{r}}}\triangleq
  \mathds{1}_{\{s \in S_{\texttt{r}}\}}$ is another indicator function.
  Moreover, their \emph{expected} transition measure
  under~$\boldsymbol{\alpha}$ are denoted
  by~$\overline{\mathcal{P}}_1$ and $\overline{\mathcal{P}}_2$,
  respectively.  \hfill $\blacksquare$
\end{definition}

\begin{theorem} \label{theorem:sat-strict}
  The probability that $\varphi$ is satisfied under belief~$\boldsymbol{\alpha}$
  and policy $\boldsymbol{\pi}_{\texttt{o}}$ at stage $0$ can be
  lower-bounded by:
\begin{equation}\label{eq:theo-sat-bound}
\begin{split}
  \mathbb{E}_{\boldsymbol{\alpha}} \{\textbf{Sat}_{\mathcal{P}}^{\boldsymbol{\pi}_{\texttt{o}}}\}
  \geq  \mathbb{E}_{\overline{\mathcal{P}}_1}^{s_0,\boldsymbol{\pi}_{\texttt{o}}}
  \sum_{t=0}^{\infty}\big{(}b_{s_t,S_{\Xi}}+\sigma_{s_t,u_t}^{\boldsymbol{\alpha}}\big{)},
\end{split}
\end{equation}
where~$b_{s_t,S_{\Xi}}$, $\overline{\mathcal{P}}_1$ are defined
in Def.~\ref{def:MDPs} and $\sigma_{s,u}^{\boldsymbol{\alpha}}\leq 0$,
$\forall (s,u)\in E$ is the \textbf{cost correction term} satisfying:
\begin{equation}\label{eq:sat-sigma}
  \sigma_{s,u}^{\boldsymbol{\alpha}}\triangleq \sum_{\check{s}\in \mathcal{K}_s^u}
  \mathbb{E}_{\boldsymbol{\alpha}}\big{\{}\min(0,\, p_{s,u}^{\check{s}}
  -\mathbb{E}_{\boldsymbol{\alpha}}\{p_{s,u}^{\check{s}}\})\big{\}},
\end{equation}
where $p_{s,u}^{\check{s}}\triangleq p_E(s,u,\check{s})$ from~\eqref{eq:new_pro}.
\end{theorem}
\begin{proof}
  It has been shown in~\cite{baier2008principles,
    etessami2007multi} that the probability
  that~$\varphi$ is satisfied under belief~$\boldsymbol{\alpha}$ equals
  to the probability that the system~$\mathcal{P}$ enters the union of
  AMECs,~i.e., $S_{\Xi}\triangleq \cup_{c=1}^C S'_c$ with $(S'_c,U'_c)\in \Xi_{acc}$.
  Thus, the left-hand side of~\eqref{eq:theo-sat-bound} can be computed by:
\begin{equation*}\label{eq:sat-bound-equal-1}
\begin{split}
  \mathbb{E}_{\boldsymbol{\alpha}}
  \{\textbf{Sat}_{\mathcal{P}}^{\boldsymbol{\pi}_{\texttt{o}}}\}
  &= \mathbb{E}_{\boldsymbol{\alpha}}
  \mathbb{E}_{\mathcal{P}}^{s_0,\boldsymbol{\pi}_{\texttt{o}}} \{B_{S_{\Xi}}\}
  = \mathbb{E}_{\boldsymbol{\alpha}}
  \mathbb{E}_{{\mathcal{P}}_1}^{s_0,\boldsymbol{\pi}_{\texttt{o}}}
  \{\sum_{t=0}^{\infty}b_{s_t,S_{\Xi}}\}, \\
\end{split}
\end{equation*}
where $B_{S_{\Xi}}=\mathds{1}_{\{\exists t<\infty, s_t\in S_{\Xi}\}}$,
$b_{s_t,S_{\Xi}}$ and ${\mathcal{P}}_1$ are defined in Def.~\ref{def:MDPs}.
Furthermore, by Lemma~3 of~\cite{moldovan2012safe}, it holds that
\begin{equation*}
\begin{split}
  \mathbb{E}_{\boldsymbol{\alpha}}
  \mathbb{E}_{{\mathcal{P}}_1}^{s_0,\boldsymbol{\pi}_{\texttt{o}}}
  \{\sum_{t=0}^{\infty}b_{s_t,S_{\Xi}}\} &=
  \mathbb{E}_{\overline{\mathcal{P}}_1}^{s_0,\boldsymbol{\pi}_{\texttt{o}}}
  \{\sum_{t=0}^{\infty}(b_{s_t,S_{\Xi}} +
  \sigma_{s,u}^{\boldsymbol{\alpha},\boldsymbol{\pi}_{\texttt{o}}})\},\\
\end{split}
\end{equation*}
where the policy-dependent correction term is given by
$\sigma_{s,u}^{\boldsymbol{\alpha},\boldsymbol{\pi}_{\texttt{o}}}\triangleq
\sum_{\check{s}\in \mathcal{K}_s^u} \mathbb{E}_{\boldsymbol{\alpha}}
\big{\{}(p_{s,u,\check{s}}
-\mathbb{E}_{\boldsymbol{\alpha}}\{p_{s,u,\check{s}}\})
\,\mathbb{E}_{\mathcal{P}_1}^{s_0,\boldsymbol{\pi}_{\texttt{o}}}
\{B_{S_{\Xi}}\}\big{\}}.$
Since $\mathbb{E}_{\mathcal{P}_1}^{s_0,\boldsymbol{\pi}_{\texttt{o}}}
\{B_{S_{\Xi}}\}\in [0, 1]$ holds for all $\boldsymbol{\pi}_{\texttt{o}}$,
we can easily show that
$\sigma_{s,u}^{\boldsymbol{\alpha},\boldsymbol{\pi}_{\texttt{o}}}\geq
\sigma_{s,u}^{\boldsymbol{\alpha}}$ holds with $\sigma_{s,u}^{\boldsymbol{\alpha}}$
defined in~\eqref{eq:sat-sigma}.
Thus, the lower bound in~\eqref{eq:theo-sat-bound}  is verified.
\end{proof}

\begin{theorem} \label{theorem:safe-strict}
  The safety constraint  under belief $\boldsymbol{\alpha}$
  for any policy $\boldsymbol{\pi}_{\texttt{o}}$
  and $\boldsymbol{\pi}_{\texttt{r}}$ at stage $0$
  can be lower-bounded by:
\begin{equation}\label{eq:theo-safe-bound}
\begin{split}
  \mathbb{E}_{\boldsymbol{\alpha}}
  \{\textbf{Safe}_{\mathcal{P}}^{\boldsymbol{\pi}_{\texttt{o}},\boldsymbol{\pi}_{\texttt{r}}}\}
  \geq \mathbb{E}_{\overline{\mathcal{P}}}^{s_0,\boldsymbol{\pi}_{\texttt{o}}}
  \big{\{}v_{s_{1}}^\star+\sigma_{s_{0},u_{0}}^{\boldsymbol{\alpha}}\big{\}},
\end{split}
\end{equation}
where the value function $v_{s}^\star \in [0, 1]$ is given by:
\begin{equation}\label{eq:safe-v}
  v^\star_{s} = \mathbb{E}_{\overline{\mathcal{P}}_2}^{s,\boldsymbol{\pi}_{\texttt{r}}}
  \big{\{}\sum_{t=0}^{\infty} \big{(} b_{s_t,S_{\texttt{r}}}
  +(1-b_{s_t,S_{\texttt{r}}})\sigma_{s_t,u_t}^{\boldsymbol{\alpha}}\big{)}\big{\}},
\end{equation}
where~$b_{s, S_{\texttt{r}}}, \overline{\mathcal{P}}_2$ are defined in
Def.~\ref{def:MDPs}, and the correction term~$\sigma_{s,u}^{\boldsymbol{\alpha}}$
is the defined the same as in~\eqref{eq:sat-sigma}.
\end{theorem}
\begin{proof}
  By the definition of safety in~\eqref{eq:return},
  the safety constraint in~\eqref{eq:objective} under
  belief $\boldsymbol{\alpha}$ can be computed by:
\begin{equation}\label{eq:safe-bound-1}
\begin{split}
  &\mathbb{E}_{\boldsymbol{\alpha}}
  \{\textbf{Safe}_{\mathcal{P}}^{\boldsymbol{\pi}_{\texttt{o}},\boldsymbol{\pi}_{\texttt{r}}}\}
  = \mathbb{E}_{\boldsymbol{\alpha}}\,
  \mathbb{E}_{\mathcal{P}}^{s_0,\boldsymbol{\pi}_{\texttt{o}}} \,
  \mathbb{E}_{\mathcal{P}}^{s_1,\boldsymbol{\pi}_{\texttt{r}}} \{B_{S_{\texttt{r}}}\} \\
  &\quad \geq \mathbb{E}_{\overline{\mathcal{P}}}^{s_0,\boldsymbol{\pi}_{\texttt{o}}}
  \big{\{}\mathbb{E}_{\boldsymbol{\alpha}}
  \mathbb{E}_{\mathcal{P}}^{s_{1},\boldsymbol{\pi}_{\texttt{r}}}\{B_{S_{\texttt{r}}}\}
  +\sigma_{s_{0},u_{0}}^{\boldsymbol{\alpha}}\big{\}},\\
\end{split}
\end{equation}
where~$B_{S_{\texttt{r}}}\triangleq
\mathds{1}_{\{\exists t<\infty, s_t\in S_{\texttt{r}}\}}$,
$\overline{\mathcal{P}}$ and $\sigma_{s,u}^{\boldsymbol{\alpha}}$ are
defined as before. The lower bound above is derived similarly as
in Theorem~\ref{theorem:sat-strict}.
Then, the inner term of~\eqref{eq:safe-bound-1} can be relaxed further
by applying again the same analysis (but for set $S_{\texttt{r}}$):
\begin{equation}\label{eq:safe-bound-2}
\begin{split}
  &\mathbb{E}_{\boldsymbol{\alpha}}\mathbb{E}_{\mathcal{P}}^{s,\boldsymbol{\pi}_{\texttt{r}}}
  \{B_{S_{\texttt{r}}}\} = \mathbb{E}_{\boldsymbol{\alpha}}
  \mathbb{E}_{\mathcal{P}_2}^{s,\boldsymbol{\pi}_{\texttt{r}}}
  \{\sum_{t=0}^{\infty} b_{s_t,S_{\texttt{r}}}\} \\
  &\quad \geq \mathbb{E}_{\overline{\mathcal{P}}_2}^{s,\boldsymbol{\pi}_{\texttt{r}}}
  \big{\{}\sum_{t=0}^{\infty} \big{(} b_{s_t,S_{\texttt{r}}}+(1-b_{s_t,S_{\texttt{r}}})
  \sigma_{s_t,u_t}^{\boldsymbol{\alpha}}\big{)}\big{\}}=v_s^\star,
\end{split}
\end{equation}
where $b_{s_t,S_{\texttt{r}}}$, ${\mathcal{P}}_2$, $\overline{\mathcal{P}}_2$
are defined in Def.~\ref{def:MDPs}.
Thus, the lower-bound of the safety constraint
in~\eqref{eq:theo-safe-bound} is verified.
\end{proof}

Note that Theorems~\ref{theorem:sat-strict} and~\ref{theorem:safe-strict}
allow us to evaluate the satisfiability and safety constraints in a
tractable way, i.e., by replacing the expectations over \emph{all}
belief of MDPs with a single MDP that has the \emph{expected} transition
measure and appropriate costs. These lower bounds would yield stricter
but tractable constraints. We now describe in the sequel how to
synthesize the control policies  using these bounds.

Lastly, since the two Dirichlet distributions by~$p_D$ and $p_L$ are
independent, the expectation of $p_E$ can be computed
analytically~\cite{mackay2003information}. Moreover, since the
marginal distribution of a Dirichlet distribution is a beta
distribution~\cite{mackay2003information}, the correction
term~$\sigma_{s,u}^{\boldsymbol{\alpha}}$ in~\eqref{eq:sat-sigma}
can be computed efficiently by Monte-Carlo estimation over each dimension.

%==========
\subsubsection{Policy Prefix Synthesis}\label{subsubsec:prefix-synthesis}
The goal of the policy prefix is to drive the system from initial state~$s_0$
to the set of AMECs $S_{\Xi}$ with minimum cost, while satisfying the
safety and satisfiability constraints. We formulate the following
constrained optimization problem:
\begin{subequations}\label{eq:prefix-optimization}
\begin{align}
  &\min_{\boldsymbol{\pi}_{\texttt{o}},\boldsymbol{\pi}_{\texttt{r}}} \;\;
  \mathbb{E}^{\boldsymbol{\pi}_{\texttt{o}}}_{\overline{\mathcal{P}}_1}
  \big{\{}\sum_{t=0}^{\infty} \big{(}c_E(s_t,u_t)-\xi_{s_t}^{u_t} \big{)}\big{\}}
  \label{eq:prefix-obj}\\
  &\text{s.t.} \quad \mathbb{E}_{\overline{\mathcal{P}}_1}^{s_0,\boldsymbol{\pi}_{\texttt{o}}}
  \big{\{}\sum_{t=0}^{\infty}(b_{s_t,S_{\Xi}}+\sigma_{s_t,u_t}^{\boldsymbol{\alpha}})\big{\}}
  \geq \chi_{\texttt{o}};  \label{eq:prefix-sat}\\
  &\qquad \mathbb{E}_{\overline{\mathcal{P}}_1}^{s_0,\boldsymbol{\pi}_{\texttt{o}}}
  \big{\{}v^\star_{s_1}+\sigma_{s_0,u_0}^{\boldsymbol{\alpha}}\big{\}}\geq
  \chi_{\texttt{r}}; \label{eq:prefix-safe-1}
\end{align}
\end{subequations}
where $\overline{\mathcal{P}}_1$, $\sigma_{s,u}^{\boldsymbol{\alpha}}$ are defined
in~\eqref{eq:theo-sat-bound}-\eqref{eq:theo-safe-bound}.
The exploration bonus $\xi_{s_t}^{u_t}$ from~\eqref{eq:bonus} is incorporated
in the objective function~\eqref{eq:prefix-obj} to encourage exploration
while minimizing the expected total cost to reach the set of AMECs $S_{\Xi}$.
The constraint~\eqref{eq:prefix-sat} ensures that the satisfiability is lower
bounded by $\chi_{\texttt{o}}$; and constraint~\eqref{eq:prefix-safe-1} ensures
that the safety is lower-bounded by $\chi_{\texttt{r}}$ with value
function~$v_s^\star$ defined in~\eqref{eq:safe-v}.
The above optimization can be solved in three steps:
First, construct~$\mathcal{P}_2$ and computes the associated value function~$v^\star_s$.
Given $v^\star_s$, problem~\eqref{eq:prefix-optimization}
can be formulated as linear programs (LP) as proposed in our earlier
work~\cite{guo2018probabilistic}.
The LP can be solved via any LP solver, based on which
the prefix of the outgoing policy can be derived.

\begin{lemma}\label{lem:prefix}
The optimal policy~$\boldsymbol{\pi}^\star_{\texttt{pre},\texttt{o}}$
derived above ensures both the reachability
constraint~${Pr}_{\mathcal{M},s_0}^{\boldsymbol{\pi}^\star_{\texttt{pre},\texttt{o}}}
(\Diamond S_{\Xi}) \geq \chi_{\texttt{o}}$ and the safety
constraint~$\mathbb{E}_{\boldsymbol{\alpha}}
\{\textbf{Safe}_{\mathcal{M}}^{\boldsymbol{\mu}_{\texttt{o}},
\boldsymbol{\mu}_{\texttt{r}}}\}\geq \chi_{\texttt{r}}$ hold.
\end{lemma}

\begin{proof}
The proof is omitted and follows directly from Theorems~\ref{theorem:sat-strict} and~\ref{theorem:safe-strict}.
\end{proof}

%====================
\subsubsection{Policy Suffix Synthesis}\label{subsubsec:suffix-synthesis}
Once the system reaches the union of AMECs $S_{\Xi}$ under the prefix
policy~$\boldsymbol{\pi}^\star_{\texttt{pre},\texttt{o}}$, the system remains
inside~$S_{\Xi}$ by following the action set given by the
AMECs~\cite{etessami2007multi}.
Thus the goal of the policy suffix is to minimize the mean total cost
defined in~\eqref{eq:total-cost} while ensuring the safety constraint.
For each AMEC $(S'_c, U'_c)\in \Xi_{acc}$,
we denote by $I'_c\triangleq S'_c\cap I^i_{\mathcal{P}}$ the goal states
that the system should intersect infinitely often,
where~$(H^i_{\mathcal{P}}, I^i_{\mathcal{P}})\in Acc_{\mathcal{P}}$ is
the associated accepting pair.
First, we construct a variant MDP of~$\mathcal{P}$ as follows.

\begin{definition}\label{def:mdp-suf}
  The MDP $\mathcal{P}_3$ is a sub-MDP of~$\mathcal{P}$ that only contains
  the states within $S_{\Xi}$ and only actions within $U_c'(s)$ are allowed,
  $\forall (S'_c, U'_c)\in \Xi_{acc}$.
\hfill $\blacksquare$
\end{definition}

Moreover,
  for each AMEC $(S'_c, U'_c)\in \Xi_{acc}$, we first split~$I_c'$ into two
  virtual copies:~$I_{\texttt{in}}$ which only has incoming transitions
  into~$I_c'$ and~$I_{\texttt{out}}$ that has only outgoing transitions
  from~$I_c'$. Once the system enters $I_{\texttt{in}}$ it remains inside
  with zero cost. Denote by $S'_d\triangleq (S'_c\backslash I'_c)
  \cup I_{\texttt{in}} \cup I_{\texttt{out}}$ the new set of states of~$\mathcal{P}_3$,
  and $S_d\triangleq S'_d \backslash I_{\texttt{in}}$.
Then, we consider the following optimization problem:
\begin{subequations}\label{eq:suffix-optimization}
\begin{align}
  &\min_{\boldsymbol{\pi}_{\texttt{o}},\boldsymbol{\pi}_{\texttt{r}}} \;\;
  \mathbb{E}^{\boldsymbol{\pi}_{\texttt{o}}}_{\overline{\mathcal{P}}_3}
  \big{\{}\lim_{t\rightarrow \infty} \frac{1}{t}\sum_{t=0}^{\infty}
  \big{(}c_E(s_t,u_t)-\xi_{s_t}^{u_t} \big{)}\big{\}}\label{eq:suffix-obj}\\
  &\text{s.t.} \quad \mathbb{E}_{\overline{\mathcal{P}}_3}^{s_0,\boldsymbol{\pi}_{\texttt{o}}}
  \big{\{}v^\star_{s_1}+\sigma_{s_0,u_0}^{\boldsymbol{\alpha}}\big{\}}\geq
  \chi_{\texttt{r}}; \label{eq:suffix-safe-1}
\end{align}
\end{subequations}
where $\overline{\mathcal{P}}_3$ is the expected measure of the
sub-MDP $\mathcal{P}_3$ defined above; $\xi_{s}^{u}$,
$\sigma_{s,u}^{\boldsymbol{\alpha}}$ and $v_s^\star$ are the computed
in the same way as in~\eqref{eq:prefix-optimization}.
The exploration bonus $\xi_{s}^{u}$ is incorporated in the
objective function~\eqref{eq:suffix-obj} to encourage exploration
while minimizing the mean total cost within the AMECs $S_{\Xi}$.
The constraint~\eqref{eq:suffix-safe-1} ensures the safety.
Note that the satisfiability constraint is not incorporated
as it is ensured by the structure of the AMEC.
Similar to the prefix, the above optimization can be solved
in three steps: construct the MDP $\overline{\mathcal{P}}_2$,
formulate and solve the LP, and synthesize the suffix of the outgoing policy.

\begin{lemma}\label{lem:suffix}
  The optimal policy suffix~$\boldsymbol{\pi}^\star_{\texttt{suf},\texttt{o}}$
  derived above minimizes the mean total cost defined
  in~\eqref{eq:total-cost} once the system has been fully explored,
  i.e., $\xi_s^u=0$, $\forall (s,u)\in E$. Moreover, the system remains
  inside $S_{\Xi}$ and the safety
  constraint~$\mathbb{E}_{\boldsymbol{\alpha}}
  \{\textbf{Safe}_{\mathcal{M}}^{\boldsymbol{\mu}_{\texttt{o}},
    \boldsymbol{\mu}_{\texttt{r}}}\}\geq \chi_{\texttt{r}}$ holds.
\end{lemma}
\begin{proof}
  First, the objective function in~\eqref{eq:suffix-optimization}
  is equivalent to the mean total cost defined
  in~\eqref{eq:total-cost} if the exploration bonus is set to zero.
  Second, due to the definition of AMECs, the system remains
  inside~$S_{\Xi}$ when the policy only chooses actions that are
  allowed by~$U'_c$. Lastly, the safety constraint is ensured by
  Theorem~\ref{theorem:safe-strict}.
\end{proof}

%==========
\subsection{Online Policy Execution and Adaptation}
\label{subsec:policy-execution}
To solve Problem~\ref{prob:main}, both the outgoing and return
policies~$\boldsymbol{\pi}_{\texttt{o}}\triangleq
\{\boldsymbol{\pi}^\star_{\texttt{pre},\texttt{o}},
\boldsymbol{\pi}^\star_{\texttt{suf},\texttt{o}}\}$
and $\boldsymbol{\pi}_{\texttt{r}}$ should be mapped
back to the policy $\boldsymbol{\mu}_{\texttt{o}}$
of $\mathcal{M}$.
First, the control policy $\mu_0$ is set
to~$\boldsymbol{\pi}_{\texttt{pre},\texttt{o}}^\star(s_0)$,
Afterwards, the robot observes the states
and updates its belief~$\boldsymbol{\alpha}$, then reaches $s_1\in S$ at stage one.
Then, $\boldsymbol{\pi}^\star_{\texttt{pre},\texttt{o}}$ is re-synthesized
but using $s_1$ as the current state and the updated~$\mathcal{P}$.
The control policy $\mu_1$ is set
to~$\boldsymbol{\pi}^\star_{\texttt{pre},\texttt{o}}(s_1)$.
This procedure repeats itself until $s_T\in S_{\Xi}$ holds at
stage~$T$ and then we switch to the policy
suffix~$\boldsymbol{\pi}^\star_{\texttt{suf},\texttt{o}}$.
Just like before, this procedure is repeated until the system is stopped.
Note that whenever the agent is requested to \emph{return} to the home
states, the return policy~$\boldsymbol{\pi}_{\texttt{r}}$ is
activated. It is mapped to the policy~$\boldsymbol{\mu}_{\texttt{r}}$
in a way similar to~$\boldsymbol{\pi}^\star_{\texttt{pre},\texttt{o}}$.

%=========================================================
\section{Case Study}\label{sec:case}
In this section, we present numerical studies in simulation.
All algorithms are implemented in Python 3.9 and tested
on a laptop (3.06GHz Duo CPU and 8GB of RAM).

\subsection{Workspace Description}\label{subsec:workspace}
A search-and-rescue ground vehicle (of size $1m\times 1m$) is
deployed to explore an large area of forest (of size $20m\times 20m$)
after a wildfire breakout. Meanwhile, it should search for injured humans
and bring them to the closest base station, while maintaining a certain
amount of water in the water tank by visiting the water reservoirs.
Note that the robot can \emph{not} visit a water reservoir
with a human victim onboard. During the whole mission,
the robot should avoid: collision into obstacles, areas of high temperature,
deep valleys that it can not escape, and high hills that it can not descend.
In particular, the properties of interest are given by: humans ($\texttt{h}$),
base stations ($\texttt{b}$), water resources ($\texttt{w}$)
and obstacle/fire areas ($\texttt{o}$).
The task described above can be specified in LTL
as~$\varphi= (\square \neg \texttt{o}) \wedge (\square
(\texttt{h} \rightarrow (\neg \texttt{w})\textsf{U} \texttt{b}))
\wedge (\square \Diamond \texttt{b}) \wedge (\square \Diamond \texttt{w})
\wedge (\square \Diamond \texttt{h})$.
The satisfiability bound $\chi_{\texttt{o}}$ is set to $0.9$.
Note that actions are omitted in the robot model, and refer the readers
to~\cite{guo2016task}.

\begin{figure}[t]
\begin{minipage}[t]{0.495\linewidth}
\centering
   \includegraphics[width =1.02\textwidth]{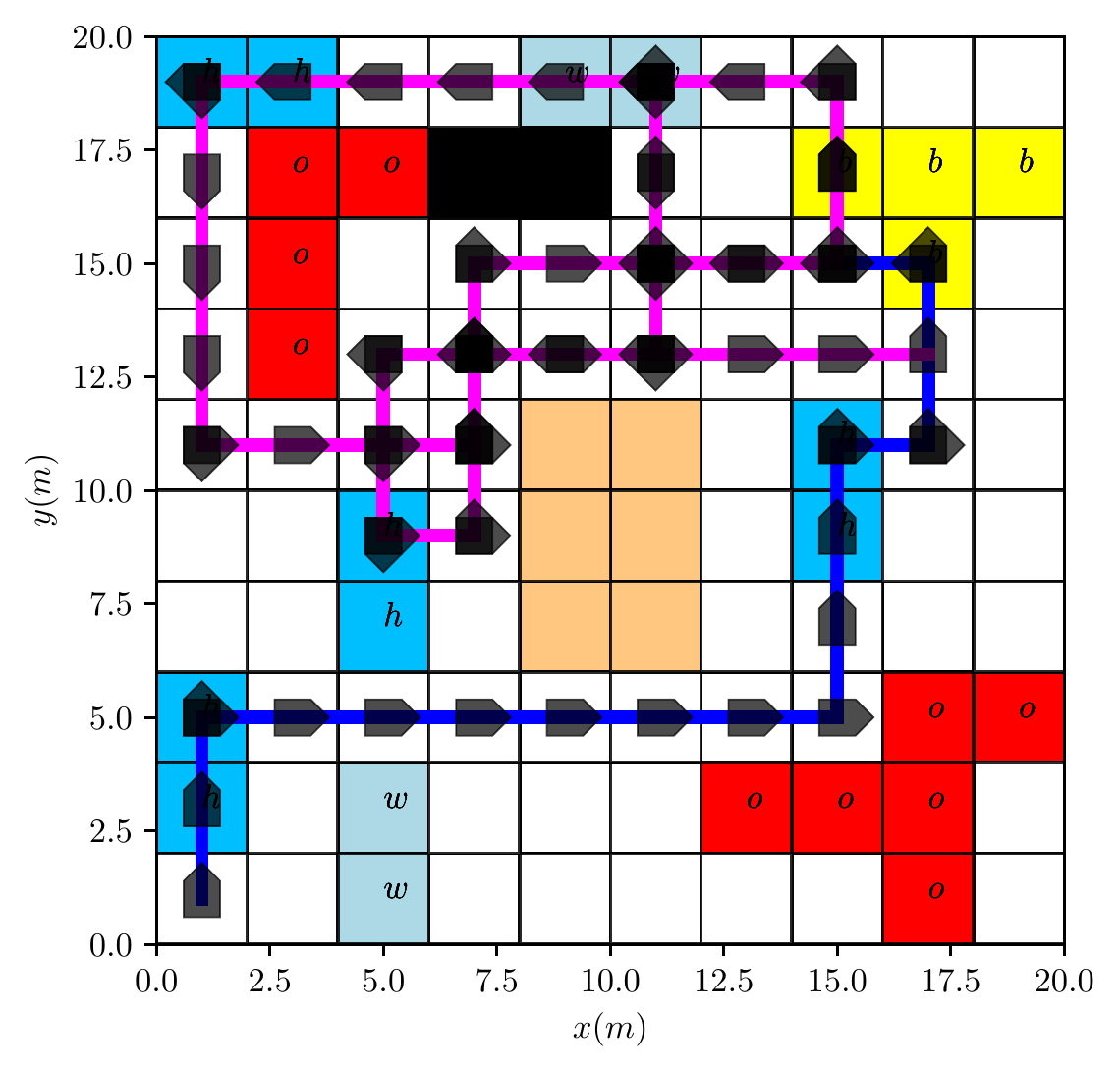}
  \end{minipage}
\begin{minipage}[t]{0.495\linewidth}
\centering
    \includegraphics[width =1.02\textwidth]{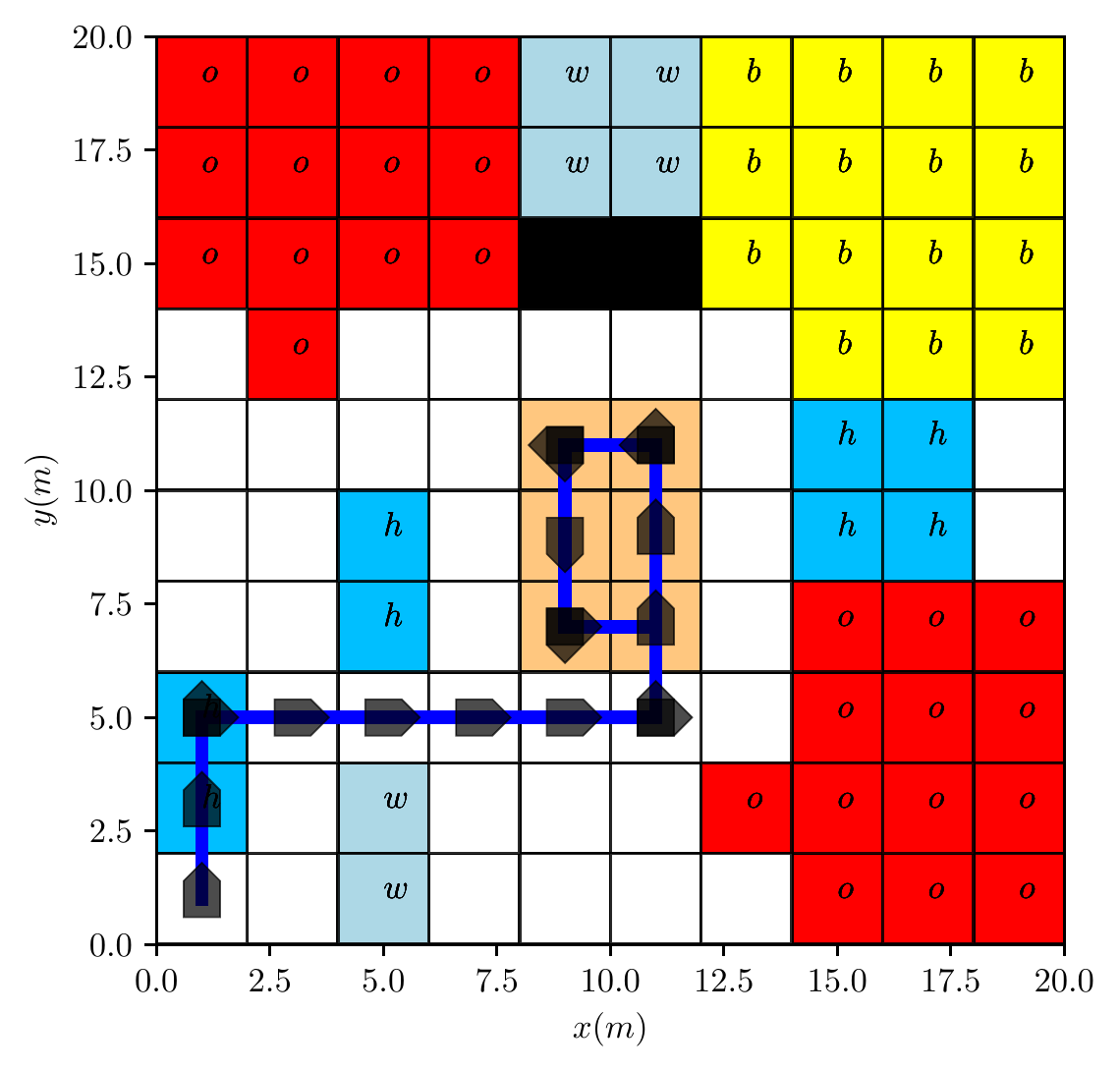}
  \end{minipage}
\caption{Sample trajectories under the proposed  policy (left)
  and under the unsafe policy (right).
  Cells are marked by the features they satisfy,
  while valleys, hills are marked in brown and black. }
\label{fig:sim-traj-1}
\end{figure}

Initial model of the forest environment (including features such as the heat map,
height map, forest density and human distribution) is obtained
from a helicopter's aerial image that has a resolution of $4m$,
which is used to construct the initial model of $\mathcal{M}$.
As shown in Fig.~\ref{fig:sim-traj-1},
the initial model provides very coarse
information about the actual workspace,
meaning that the robot would need to explore the workspace actively.
Moreover, the robot is equipped with sensors to measure the features
mentioned above within a $6m\times 6m$ area around it,
however with a increasing uncertainty by distance ($10\%$ every $2m$).
The partitioned cells are of size $2m\times 2m$ and
the robot can only move to the adjacent cells via actions:
forward, left, right, backward (with cost $3$, $5$, $5$, $6$).
It can ascend a hill of maximum angle $15^{\circ}$
and descend a slope of maximum $20^{\circ}$.
The home state is set to the robot's initial state and
the safety bound $\chi_{\texttt{r}}$ is set to $0.8$.

%========================================
%%%%%%%%%%%%%%%%%%%%

\subsection{Simulation Results}\label{subsec:results}

The underlying MDP~$\mathcal{M}$ has $400$ states and $3616$ edges
and the DRA~$\mathcal{A}_{\varphi}$ contains $21$ states, $111$ edges
and one accepting pair. It took $5.7s$ to construct the resulting
product~$\mathcal{P}$ which has $8400$ states, $75936$ edges and
one AMECs. We follow the policy synthesis and execution described
in Sec.~\ref{sec:synthesis}. When the system starts,
the robot's state is outside the AMEC.
It took $0.01s$ to calculate the value function $v_{s}^\star$
via linear program~\cite{davis2018markov} for the return policy.
Then we formulate and solve~\eqref{eq:prefix-optimization} given $v_{s}^\star$
for the prefix synthesis in $0.02s$,
which contains $2840$ variables and $712$ constraints.
An optimal action is chosen based on plan prefix.
Then the robot takes new measurements
and updates~$\mathcal{M}$ and~$\mathcal{P}$ in a Bayesian way,
which takes in average~$1.5s$.
This process repeats itself until the robot reaches the set of AMECs.
Then the optimization~\eqref{eq:suffix-optimization} for
the plan suffix is formulated (with $598$ variables and $1484$ constraints)
and solved within $0.25s$.
One sample trajectory is shown in Fig.~\ref{fig:sim-traj-1},
which satisfies the assigned search and rescue task.
The trajectory prefix is marked in blue while the suffix is marked in magenta.
It can also be seen that after the exploration and learning,
the final workspace model is the same as the actual model.
More importantly, due to the enforced safety constraint,
the robot avoids during exploration  the area of deep valley (in brown)
and high hills (in black). In comparison, we also simulate
the robot trajectory under the same synthesis algorithm
but removing the safety constraints.
One sample trajectory is shown in Fig.~\ref{fig:sim-traj-1}
where the robot remains trapped in the valley after time~$9$.
Thus it fails to satisfy the formula and leaves most of the workspace unexplored.
%========================================

\subsection{Performance Evaluation}\label{subsec:evaluation}

In order to further demonstrate the computational complexity of the
proposed approach, we run $100$ Monte Carlo simulations of the above
robotic system under different sizes of the underlying map,
with similar setup of features.
In Table~\ref{table:compare}, we record the average synthesis time,
the task satisfiability and the safety measure,
which are compared with the approach that does not consider safety
during exploration. First, it can be noticed that the proposed algorithm
scales well with workspace size (with millions of edges in the last case).
Constructing the product $\mathcal{P}$ takes considerable amount of time
while solving the LPs associated with~\eqref{eq:prefix-optimization}
and~\eqref{eq:suffix-optimization} are relatively fast.
Second, it can be seen that both the task satisfiability
and safety measure are greatly improved under the proposed approach.
This is because in partially-known workspaces violating the safety
constraint would also indicate the violation of assigned temporal task.

%%%%%%%%%
\begin{table}[t]
\begin{center}
{\renewcommand{\arraystretch}{1.4}
\scalebox{1.05}{
    \begin{tabular}{|c|c|c|c|c|c|c|}
    \hline
\textbf{Approach} & $\mathcal{P}$ Size & \multicolumn{2}{c|}{Time[$s$]}  & Safety & Satisfy \\
 \hhline{|======|}
 \multirow{3}{*}{Proposed} & (8.4\texttt{e}3,\, 7.6\texttt{e}4) & 5.7 & 0.03 & 0.87 & 0.92 \\
\cline{2-6}
 & (2.2\texttt{e}4,\, 2.1\texttt{e}5) & 34 & 0.42  & 0.91 & 0.95 \\
\cline{2-6}
 & (1.4\texttt{e}5,\, 1.4\texttt{e}6) & 210 & 7.8  & 0.93 & 0.97 \\
 \hhline{|======|}
 \multirow{2}{*}{Unsafe} & (8.4\texttt{e}3,\, 7.6\texttt{e}4) & 5.7 & 0.01 & 0.1 & 0.3 \\
\cline{2-6}
 & (2.2\texttt{e}4,\, 2.1\texttt{e}5) & 30 & 0.31  & 0.2 & 0.25 \\
\hline
    \end{tabular}
}
}
\caption{Comparison of complexity and performance
  between the proposed and the unsafe approach.
  The ``Time'' column is split into the time to construct $\mathcal{P}$
  and to synthesize $\boldsymbol{\pi}_{\texttt{o}}$.
  Note~$a\texttt{e}b\triangleq a\times 10^b$.}
\label{table:compare}
\end{center}
\end{table}
%%%%%%%%%

%=========================================================
\section{Summary and Future Work}\label{sec:future}

This work proposes a planning framework
for robots operating in uncertain environments.
The robotic task is specified as LTL formulas.
During the learning and exploration,
we enforce a safety constraint as the probability of
returning to a set of home states during run time.
The proposed approach fulfill
both the temporal task and safety constraints.
Future work includes the consideration of multi-robot systems.
%==============================
\bibliography{meng.bib}

\begin{thebibliography}{10}
\providecommand{\url}[1]{#1}
\csname url@samestyle\endcsname
\providecommand{\newblock}{\relax}
\providecommand{\bibinfo}[2]{#2}
\providecommand{\BIBentrySTDinterwordspacing}{\spaceskip=0pt\relax}
\providecommand{\BIBentryALTinterwordstretchfactor}{4}
\providecommand{\BIBentryALTinterwordspacing}{\spaceskip=\fontdimen2\font plus
\BIBentryALTinterwordstretchfactor\fontdimen3\font minus
  \fontdimen4\font\relax}
\providecommand{\BIBforeignlanguage}[2]{{%
\expandafter\ifx\csname l@#1\endcsname\relax
\typeout{** WARNING: IEEEtran.bst: No hyphenation pattern has been}%
\typeout{** loaded for the language `#1'. Using the pattern for}%
\typeout{** the default language instead.}%
\else
\language=\csname l@#1\endcsname
\fi
#2}}
\providecommand{\BIBdecl}{\relax}
\BIBdecl

\bibitem{davis2018markov}
M.~H. Davis, \emph{Markov models and optimization}.\hskip 1em plus 0.5em minus
  0.4em\relax Routledge, 2018.

\bibitem{ding2014optimal}
X.~Ding, S.~L. Smith, C.~Belta, and D.~Rus, ``Optimal control of markov
  decision processes with linear temporal logic constraints,'' \emph{Automatic
  Control, IEEE Transactions on}, vol.~59, no.~5, pp. 1244--1257, 2014.

\bibitem{etessami2007multi}
K.~Etessami, M.~Kwiatkowska, M.~Y. Vardi, and M.~Yannakakis, ``Multi-objective
  model checking of markov decision processes,'' in \emph{Tools and Algorithms
  for the Construction and Analysis of Systems}.\hskip 1em plus 0.5em minus
  0.4em\relax Springer, 2007, pp. 50--65.

\bibitem{baier2008principles}
C.~Baier and J.-P. Katoen, \emph{Principles of model checking}.\hskip 1em plus
  0.5em minus 0.4em\relax MIT press Cambridge, 2008.

\bibitem{bruyere2016meet}
V.~Bruyere, E.~Filiot, M.~Randour, and J.-F. Raskin, ``Meet your expectations
  with guarantees: Beyond worst-case synthesis in quantitative games,''
  \emph{Information and Computation}, 2016.

\bibitem{guo2018probabilistic}
M.~Guo and M.~M. Zavlanos, ``Probabilistic motion planning under temporal tasks
  and soft constraints,'' \emph{IEEE Transactions on Automatic Control},
  vol.~63, no.~12, pp. 4051--4066, 2018.

\bibitem{kantaros2022perception}
Y.~Kantaros, S.~Kalluraya, Q.~Jin, and G.~J. Pappas, ``Perception-based
  temporal logic planning in uncertain semantic maps,'' \emph{IEEE Transactions
  on Robotics}, 2022.

\bibitem{ding2014ltl}
X.~Ding, M.~Lazar, and C.~Belta, ``{LTL} receding horizon control for finite
  deterministic systems,'' \emph{Automatica}, vol.~50, no.~2, pp. 399--408,
  2014.

\bibitem{ahmed2017sampling}
A.~Ahmed, P.~Varakantham, M.~Lowalekar, Y.~Adulyasak, and P.~Jaillet,
  ``Sampling based approaches for minimizing regret in uncertain markov
  decision processes (mdps),'' \emph{Journal of Artificial Intelligence
  Research}, vol.~59, pp. 229--264, 2017.

\bibitem{guo2015multi}
M.~Guo and D.~V. Dimarogonas, ``Multi-agent plan reconfiguration under local
  {LTL} specifications,'' \emph{The International Journal of Robotics
  Research}, vol.~34, no.~2, pp. 218--235, 2015.

\bibitem{qian2019exploration}
J.~Qian, R.~Fruit, M.~Pirotta, and A.~Lazaric, ``Exploration bonus for regret
  minimization in discrete and continuous average reward mdps,'' \emph{Advances
  in Neural Information Processing Systems}, vol.~32, 2019.

\bibitem{talebi2018variance}
M.~S. Talebi and O.-A. Maillard, ``Variance-aware regret bounds for
  undiscounted reinforcement learning in mdps,'' in \emph{Algorithmic Learning
  Theory}.\hskip 1em plus 0.5em minus 0.4em\relax PMLR, 2018, pp. 770--805.

\bibitem{garcia2015comprehensive}
J.~Garc{\i}a and F.~Fern{\'a}ndez, ``A comprehensive survey on safe
  reinforcement learning,'' \emph{Journal of Machine Learning Research},
  vol.~16, no.~1, pp. 1437--1480, 2015.

\bibitem{klein2007ltl2dstar}
J.~Klein, ``ltl2dstar-{LTL} to deterministic streett and rabin automata,''
  \texttt{\url{http://www.ltl2dstar.de}}, 2007.

\bibitem{mackay2003information}
D.~J. MacKay, \emph{Information theory, inference and learning
  algorithms}.\hskip 1em plus 0.5em minus 0.4em\relax Cambridge university
  press, 2003.

\bibitem{neal2000markov}
R.~M. Neal, ``Markov chain sampling methods for dirichlet process mixture
  models,'' \emph{Journal of computational and graphical statistics}, vol.~9,
  no.~2, pp. 249--265, 2000.

\bibitem{moldovan2012safe}
T.~M. Moldovan and P.~Abbeel, ``Safe exploration in markov decision
  processes,'' in \emph{29th International Coference on Machine
  Learning}.\hskip 1em plus 0.5em minus 0.4em\relax ACM, 2012, pp. 1451--1458.

\bibitem{guo2016task}
M.~Guo and D.~V. Dimarogonas, ``Task and motion coordination for heterogeneous
  multiagent systems with loosely coupled local tasks,'' \emph{IEEE
  Transactions on Automation Science and Engineering}, vol.~14, no.~2, pp.
  797--808, 2016.

\end{thebibliography}

\end{document}